\documentclass[lettersize,journal]{IEEEtran}
\usepackage{amsmath,amsfonts}
\usepackage{algorithmic}
\usepackage{algorithm}
\usepackage{array}
\usepackage{textcomp}
\usepackage{stfloats}
\usepackage{url}
\usepackage{verbatim}
\usepackage{graphicx}
\usepackage{cite}
\usepackage{microtype}      % microtypography
\usepackage{xcolor}         % colors
\usepackage{caption}
\usepackage{booktabs}
\usepackage{amssymb}
\usepackage{pifont}
\usepackage{subcaption}
\usepackage{multirow}
\usepackage{xspace}
\usepackage{dsfont}
\usepackage{enumitem}
\usepackage{listings}
\usepackage{scalerel}
\usepackage{soul}
\usepackage{longtable}
\usepackage{wrapfig}

\makeatletter
\DeclareRobustCommand\onedot{\futurelet\@let@token\@onedot}
\def\@onedot{\ifx\@let@token.\else.\null\fi\xspace}

\def\eg{\emph{e.g}\onedot} 
\def\ie{\emph{i.e}\onedot}

\makeatother

\newlength\savewidth
\newcommand{\tablestyle}[2]{\setlength{\tabcolsep}{#1}\renewcommand{\arraystretch}{#2}\centering\footnotesize}
\renewcommand{\paragraph}[1]{\vspace{1.25mm}\noindent\textbf{#1}}

\usepackage{array}
\newcolumntype{x}[1]{>{\centering\arraybackslash}p{#1pt}}
\newcolumntype{y}[1]{>{\raggedright\arraybackslash}p{#1pt}}
\newcolumntype{z}[1]{>{\raggedleft\arraybackslash}p{#1pt}}

\begin{document}

\title{From Semantics to Pixels: Coarse-to-Fine Masked Autoencoders for Hierarchical Visual Understanding}

\author{Wenzhao Xiang, Yue Wu, Hongyang Yu, Feng Gao, Fan Yang, Xilin Chen
        % <-this % stops a space
\thanks{Wenzhao Xiang (xiangwenzhao22@mails.ucas.ac.cn), Yue Wu (wuyue221@mails.ucas.ac.cn), and Xilin Chen (xlchen@ict.ac.cn) are with the Key Laboratory of Intelligent Information Processing, Institute of Computing Technology, Chinese Academy of Sciences, Beijing 100190, China, and also with the University of Chinese Academy of Sciences, Beijing 100049, China. Wenzhao Xiang, Yue Wu, and Hongyang Yu (yuhy01@pcl.ac.cn) are with Pengcheng Laboratory, Shenzhen 518108, China. Feng Gao (gaof@pku.edu.cn) and Fan Yang (fyang.eecs@pku.edu.cn) are with the School of Arts, Peking University, Beijing 100871, China.}% <-this % stops a space
\thanks{This work has been submitted to the IEEE for possible publication.}}

\markboth{Under Review}%
{Xiang \MakeLowercase{\textit{et al.}}: From Semantics to Pixels: Coarse-to-Fine Masked Autoencoders}


\maketitle

\begin{abstract}
Self-supervised visual pre-training methods face an inherent tension: contrastive learning (CL) captures global semantics but loses fine-grained detail, while masked image modeling (MIM) preserves local textures but suffers from ``attention drift'' due to semantically-agnostic random masking. We propose C2FMAE, a coarse-to-fine masked autoencoder that resolves this tension by explicitly learning hierarchical visual representations across three data granularities: semantic masks (scene-level), instance masks (object-level), and RGB images (pixel-level). Two synergistic innovations enforce a strict top-down learning principle. First, a \textbf{cascaded decoder} sequentially reconstructs from scene semantics to object instances to pixel details, establishing explicit cross-granularity dependencies that parallel decoders cannot capture. Second, a \textbf{progressive masking} curriculum dynamically shifts the training focus from semantic-guided to instance-guided and finally to random masking, creating a structured learning path from global context to local features. To support this framework, we construct a large-scale multi-granular dataset with high-quality pseudo-labels for all 1.28M ImageNet-1K images. Extensive experiments show that C2FMAE achieves significant performance gains on image classification, object detection, and semantic segmentation, validating the effectiveness of our hierarchical design in learning more robust and generalizable representations.

\end{abstract}


\section{Introduction}
Among the diverse paradigms in computer vision pre-training, contrastive-based~\cite{he2020momentum,chen2020simple,caron2021emerging} and reconstruction-based~\cite{bao2021beit, he2022masked,xie2022simmim} self-supervised learning have been particularly influential. Despite their tremendous success, both paradigms exhibit inherent, almost opposing, limitations that curtail their ability to learn truly comprehensive and universal visual representations.

\begin{figure}[!t]
  \centering
  \includegraphics[width=\linewidth]{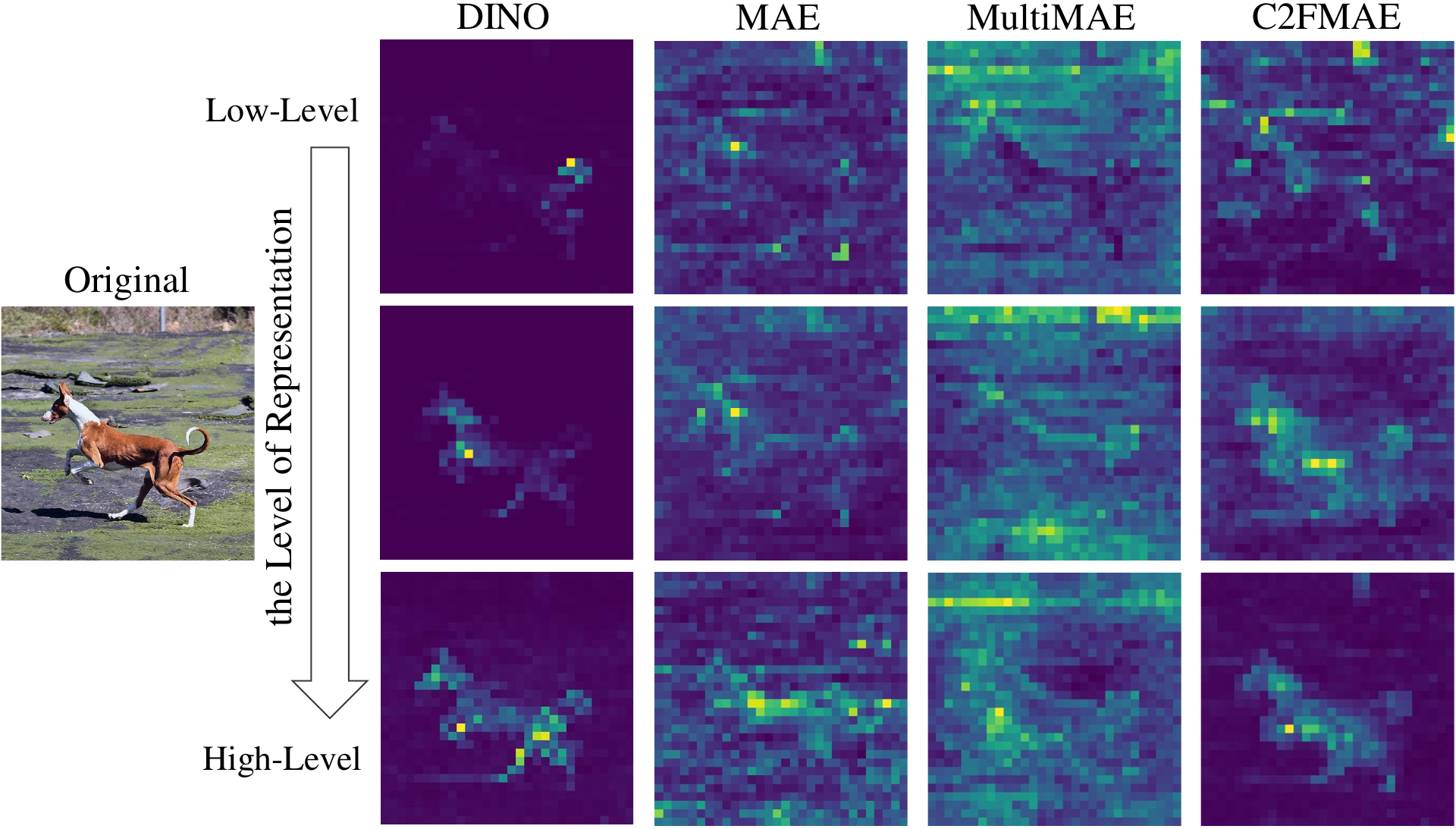}
  \caption{
  Attention maps from different methods, highlighting their representational focus. DINO excels at capturing high-level semantics, while MAE and MultiMAE's attention is directed toward low-level features. In contrast, our C2FMAE effectively captures features across all levels, successfully building a more robust hierarchical representation.
  }
   \label{fig:attention_map}
\end{figure}

Contrastive learning (CL), which pulls together global features from different views of an image, excels at learning high-level semantic representations. This makes it highly effective for image-level tasks like classification and produces clean, object-centric attention maps as shown for DINO~\cite{caron2021emerging} in Figure~\ref{fig:attention_map}. However, this strong focus on high-level semantics inherently limits the capture of fine-grained spatial information. The loss of low-level detail can limit its performance on dense prediction tasks that require precise localization and texture understanding, such as object detection and semantic segmentation~\cite{zhang2020rethinking,mahajan2018exploring}.

In contrast, Masked Image Modeling (MIM) has emerged to learn rich spatial information by reconstructing masked patches. While MIM's pixel-level objective preserves fine-grained details, its semantically-agnostic random masking strategy struggles to guide the model toward semantically critical regions~\cite{kakogeorgiou2022hide,li2022semmae}. The model often allocates significant representational capacity to reconstructing simple, low-level areas, while only crudely modeling the core objects of interest. As shown in Figure~\ref{fig:attention_map}, MAE~\cite{he2022masked} and MultiMAE~\cite{bachmann2022multimae} produce diffuse attention maps that fail to focus on salient objects.

We refer to this broader phenomenon as ``attention drift'': existing pre-training paradigms each develop a biased focus toward certain representational levels, where CL drifts toward high-level semantics while MIM drifts toward low-level textures, thus both failing to learn a complete, hierarchical understanding of the visual world.
An intuitive solution is to introduce semantic guidance to focus the reconstruction on salient foreground objects, a strategy shown to improve downstream performance~\cite{li2022semmae,sick2025attention}. However, existing methods often rely on inaccurate, self-generated attention maps for this guidance~\cite{kakogeorgiou2022hide,li2022semmae}. While modern segmentation models~\cite{kirillov2023segment,ren2024grounded} can provide far more precise masks, we argue that simply using them to distinguish foreground from background offers only a rudimentary, binary form of semantic guidance that is insufficient to resolve attention drift.

A truly comprehensive visual understanding is inherently multi-granular, requiring a model to perceive the world at multiple levels of abstraction simultaneously, ranging from coarse scene layouts to intermediate object instances and fine-grained pixel details. This hierarchical, coarse-to-fine principle is not only a long-standing and effective strategy in computer vision~\cite{lin2017feature,jiang2022coarse} but is also deeply rooted in the efficient processing pipeline of biological vision~\cite{navon1977forest,serre2014hierarchical}, offering proven advantages in learning speed and generalization~\cite{cho2021rethinking,chen2023cf}. An ideal pre-training framework should therefore unify the high-level semantic understanding of CL with the fine-grained detail preservation of MIM through explicit hierarchical guidance.

To this end, we propose C2FMAE, a vision pre-training framework that deeply integrates the coarse-to-fine principle into masked autoencoding. Our framework incorporates three visual modalities of different granularities: semantic segmentation masks (scene-level), instance segmentation masks (object-level), and RGB images (pixel-level). We enforce the coarse-to-fine principle through two synergistic innovations. First, we design a \textbf{cascaded decoder}, as opposed to a traditional parallel structure~\cite{bachmann2022multimae}. It first predicts scene-level semantic masks, then object-level instance masks, and finally reconstructs pixel-level RGB images. This pipeline ensures that the feature refinement process strictly follows a top-down path from high-level abstractions to low-level details. Second, to combat the ``attention drift'' caused by random masking, we design a \textbf{progressive masking strategy} that aligns with the objectives of our cascaded decoder. During training, the masking focus follows a carefully designed curriculum, smoothly transitioning from semantic guidance (focusing on scene regions) to instance guidance (focusing on objects), and finally to random masking (focusing on local details).

To support our framework, we construct a large-scale multi-granular dataset by generating high-quality aligned instance and semantic segmentation pseudo-labels for all 1.28M images in ImageNet-1K. Through the deep synergy of the cascaded decoder and progressive masking, C2FMAE embeds the coarse-to-fine principle into every stage of pre-training. This effectively overcomes the limitations of previous paradigms, enabling the model to learn more robust and generalizable hierarchical visual representations. As visually evidenced in Figure~\ref{fig:attention_map}, our method produces attention maps that perform well across different representation levels, thereby resolving the ``attention drift" issue and validating the successful construction of a true hierarchical visual representation. 

Our key contributions are summarized as follows:
\begin{itemize}[leftmargin=2em]
    \item We propose C2FMAE, a coarse-to-fine vision pre-training framework that learns hierarchical visual representations by jointly leveraging RGB images, instance masks, and semantic masks across three levels of granularity.
    \item We introduce two synergistic innovations: a \textbf{cascaded decoder} that sequentially refines features from scene-level semantics to pixel-level details, and a \textbf{progressive masking strategy} that dynamically shifts the training focus to align with this top-down hierarchy.
    \item We construct a large-scale multi-granular dataset by generating high-quality aligned instance and semantic segmentation pseudo-labels for all 1.28M images in ImageNet-1K. Beyond supporting our framework, this dataset serves as a valuable public resource for the broader vision community, facilitating future research in areas such as multi-modal foundation models, weakly-supervised dense prediction, and layout-controllable image generation.
    \item Extensive experiments show C2FMAE achieves significant performance gains across multiple vision tasks, such as image classification, object detection, and semantic segmentation, with superior training efficiency.
\end{itemize}

\begin{figure*}[t]
  \centering
   \includegraphics[width=\linewidth]{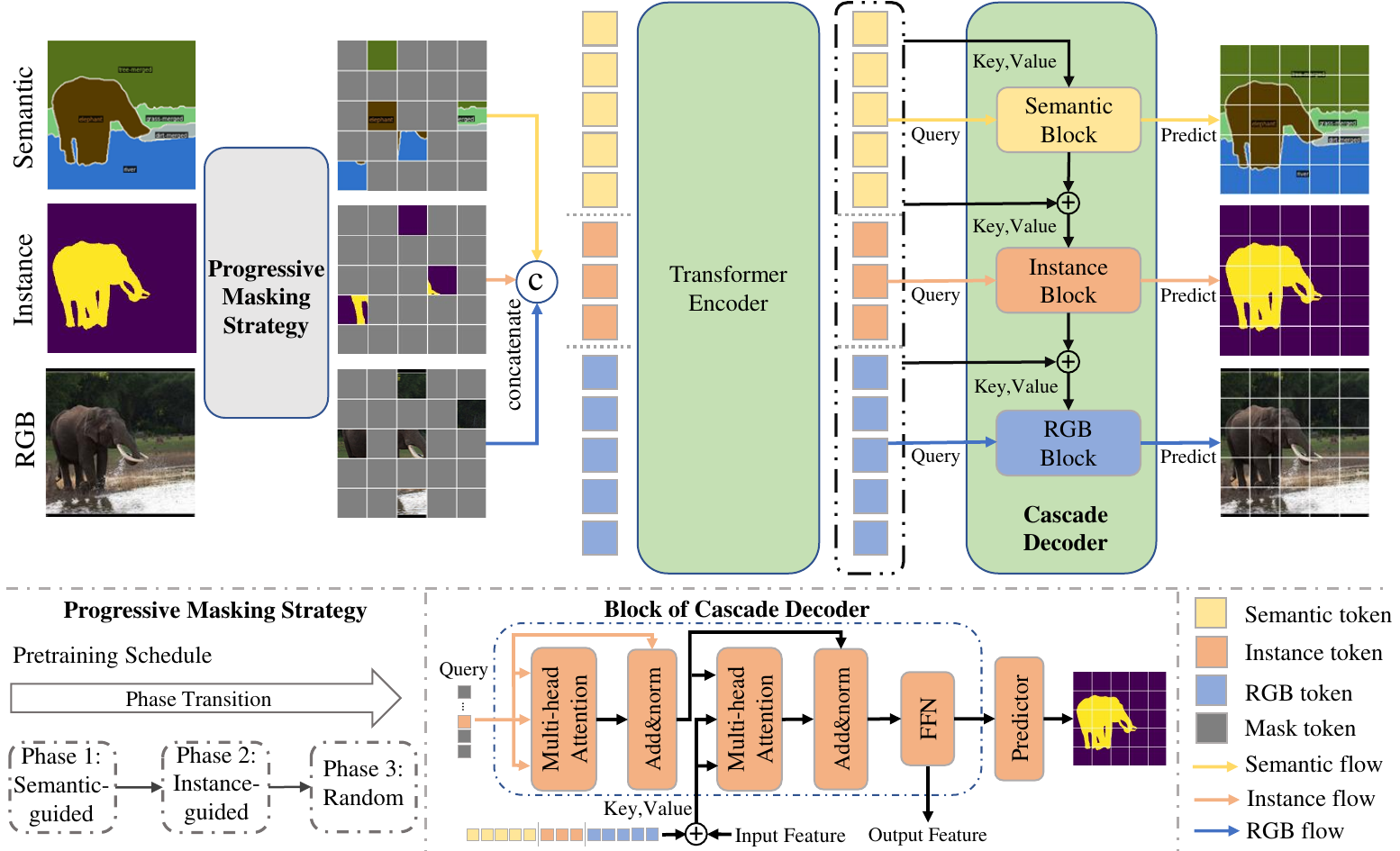}
   \caption{C2FMAE pre-training framework. Multi-granular data (RGB, Instance, Semantic masks) is first masked by Progressive Masking and then concatenated, and fed to a transformer encoder. Encoded tokens subsequently flow into a cascaded decoder with three task-specific blocks. Each block is a standard Transformer decoder block, composed of self-attention, cross-attention, and feed-forward network layers. We use linear layers as the final predictor. As training progresses, the masking strategy transitions from semantic-guided masking to instance-guided masking, and finally to random masking to build hierarchical visual representations.}
   \label{fig:framework}
\end{figure*}

\section{Related Work}
\label{sec:related_works}
\subsection{Masked Image Modeling}

Masked Image Modeling (MIM) has evolved rapidly as a dominant paradigm in self-supervised vision learning. We categorize existing approaches into three streams: foundational reconstruction frameworks, structured masking strategies, and multi-modal pre-training.

\noindent\textbf{Foundational MIM Frameworks.}
Inspired by BERT in NLP, BEiT~\cite{bao2021beit} pioneered the prediction of discrete visual tokens from masked image patches. Following this, MAE~\cite{he2022masked} established a scalable baseline by reconstructing raw pixels using an asymmetric encoder-decoder with a high masking ratio (75\%). This prompted a wave of architectural refinements: SimMIM~\cite{xie2022simmim} demonstrated that a lightweight linear decoder suffices for fine-tuning performance; iBOT~\cite{zhou2022image} combined MIM with contrastive learning through an online tokenizer; and MaskFeat~\cite{wei2022masked} proposed reconstructing HOG features to focus on structural information. Other works have explored alternative reconstruction targets, such as high-level deep features~\cite{wang2023masked, ren2023deepmim}, frequency components~\cite{xie2022masked,xiang2025wavelet}, or integrated adversarial examples~\cite{xiang2026aemim}. While these methods excel at learning local patterns, their reliance on random masking often leads to weak semantic awareness, a limitation our work addresses through explicit semantic guidance.

\noindent\textbf{Structured and Semantic Masking.}
To mitigate the semantic agnosticism of random masking, several works have introduced structured masking strategies. MST~\cite{li2021mst} and AttMask~\cite{kakogeorgiou2022hide} leverage attention maps from teacher networks to guide masking toward salient regions. ADIOS~\cite{shi2022adversarial} and AutoMAE~\cite{chen2023improving} employ adversarial games to learn harder masks dynamically. More closely related to our work, SemMAE~\cite{li2022semmae} utilizes semantic parts for masking, and UnMAE~\cite{li2022uniform} introduces broad uniform masking. However, these methods typically employ a static or single-stage strategy with a fixed inductive bias. In contrast, our C2FMAE introduces a \textit{progressive} masking curriculum that dynamically evolves from semantic-focus to instance-focus and finally to random masking, mimicking a human-like learning process.

\noindent\textbf{Multi-modal MIM.}
Recent advancements have extended MIM to multi-modal settings. MultiMAE~\cite{bachmann2022multimae} is a representative work that reconstructs RGB, depth, and semantic segmentation maps simultaneously. However, MultiMAE employs a \textit{parallel} decoder architecture where different modalities are treated as independent tasks sharing a latent representation. We argue this design ignores the inherent hierarchical dependency between modalities, where scene semantics should ideally guide object definition, which in turn confines textural details. Our C2FMAE addresses this by replacing the parallel structure with a \textit{cascaded} decoder that explicitly enforces a coarse-to-fine information flow.

\subsection{Hierarchical Visual Representation Learning}
The principle of processing visual information hierarchically, from coarse global structures to fine local details, is deeply rooted in both biological vision~\cite{navon1977forest,serre2014hierarchical} and classic computer vision~\cite{lin2017feature}.

\noindent\textbf{Hierarchical Architectures.}
Many modern backbones enforce hierarchy through architectural constraints. The Swin Transformer~\cite{liu2021swin} reintroduced hierarchical feature maps to Vision Transformers using shifted windows. Similarly, H-ViT~\cite{ghahremani2024h} and Hiera~\cite{ryali2023hiera} design specialized multi-scale attention mechanisms to capture features at different resolutions. While effective, these methods bake hierarchy into the \textit{architecture}. Our goal differs: we aim to imprint hierarchical reasoning into a standard plain Vision Transformer (ViT) through the \textit{pre-training objective} itself, making the learned representation inherently hierarchical regardless of the backbone architecture.

\noindent\textbf{Coarse-to-Fine Learning Strategies.}
The coarse-to-fine strategy has been successfully applied to specific downstream tasks, such as accelerating inference in CF-ViT~\cite{chen2023cf} or refining segmentation in MIMO-UNet~\cite{cho2021rethinking}. In the realm of self-supervised learning, approaches like HGCLIP~\cite{xia2023hgclip} align images with hierarchical text descriptions, and \cite{zhang2020self} utilizes iterative grouping for unsupervised segmentation. However, a unified pre-training framework that explicitly models the generative process from semantics to pixels has been missing. C2FMAE fills this gap by integrating multi-granular data (Semantic $\to$ Instance $\to$ RGB) into a unified masked autoencoding framework, establishing a new paradigm for learning robust hierarchical representations.

\section{Method}
\label{sec:method}

This section details the proposed C2FMAE framework. We first describe the construction of our multi-granular dataset in Section~\ref{sec:3.1}, which provides the foundation for hierarchical learning. As illustrated in Figure~\ref{fig:framework}, the C2FMAE architecture consists of three core components: a shared encoder that processes multi-granular inputs into a unified representation (Section~\ref{sec:encoder}), a cascaded decoder that sequentially reconstructs from coarse semantics to fine-grained pixels (Section~\ref{sec:cascaded_decoder}), and a progressive masking strategy that dynamically guides the training curriculum (Section~\ref{sec:masking_strategy}). The training objectives are described in Section~\ref{sec:loss}.

\subsection{Dataset Construction}
\label{sec:3.1}

A prerequisite for our coarse-to-fine pre-training is a dataset with aligned annotations across different levels of granularity. To this end, we construct a large-scale multi-granular dataset by augmenting the complete ImageNet-1K training set (1.28M images) with precisely aligned instance-level and semantic-level segmentation masks.

\noindent\textbf{Instance Segmentation Masks.}
We develop a two-stage pipeline based on Grounded SAM~\cite{ren2024grounded}. In the first stage, Grounded DINO~\cite{liu2023grounding} detects all relevant objects using category names as text prompts. In the second stage, the detected regions are processed by HQ-SAM~\cite{ke2024segment}, which significantly refines mask quality, particularly around object boundaries. To balance coverage and precision, we adopt an adaptive confidence thresholding strategy: starting with a high confidence threshold, we generate initial masks for images with clearly visible objects, and then gradually lower the threshold for images with few or no detections. This prevents the generation of excessive irrelevant masks while ensuring comprehensive annotation coverage.

\noindent\textbf{Semantic Segmentation Masks.}
We employ the large-size SEEM model~\cite{zou2024segment} to generate semantic segmentation annotations. Compared to the Mask2Former~\cite{cheng2022masked} with Swin-S~\cite{liu2021swin} backbone used in MultiMAE~\cite{bachmann2022multimae}, our SAM-based approach provides superior boundary accuracy and stronger generalization across the diverse visual scenarios in ImageNet. Following the COCO~\cite{lin2014microsoft} categorical structure, we define 133 classes comprising 80 thing classes for countable objects and 53 stuff classes for uncountable background elements, enabling the model to capture both foreground objects and contextual information.

Representative samples from our dataset are visualized in Appendix, where each cell presents the original RGB image alongside its corresponding instance and semantic masks, illustrating the segmentation precision across diverse categories.

\subsection{Multi-granular Inputs and Shared Encoder}
\label{sec:encoder}

Our framework utilizes data at three distinct granularities: RGB images ($\mathbf{I}_{rgb} \in \mathbb{R}^{H \times W \times 3}$), instance segmentation masks ($\mathbf{I}_{ins} \in \mathbb{Z}^{H \times W}$), and semantic segmentation masks ($\mathbf{I}_{sem} \in \mathbb{Z}^{H \times W}$), where $H, W$ are the image dimensions.
Each granularity of input $\mathbf{I}_m$ for $m \in \{rgb, ins, sem\}$ is first divided into a sequence of $N$ non-overlapping patches. These patches are then flattened and mapped to $D$-dimensional embeddings through granularity-specific linear projection layers, resulting in a sequence of tokens $\mathbf{Z}_m = \{\mathbf{z}_m^1, \dots, \mathbf{z}_m^N\} \in \mathbb{R}^{N \times D}$.
Following the application of our progressive masking strategy (detailed in Section~\ref{sec:masking_strategy}), which yields a binary mask, only the visible tokens of each granularity are selected. The visible tokens across all granularities are then concatenated into a single sequence $\mathbf{Z}^{vis}$ and fed into a shared Vision Transformer~\cite{dosovitskiy2021an} encoder:
\begin{equation}
  \mathbf{H}_{enc} = \text{Encoder}(\mathbf{Z}^{vis}),
\end{equation}
where $\mathbf{H}_{enc} \in \mathbb{R}^{N_{vis} \times D}$ is the sequence of encoded visible tokens, with $N_{vis}$ being the number of visible tokens. This unified representation serves as a comprehensive foundation for the subsequent hierarchical feature refinement in the cascaded decoder.

\subsection{Cascaded Decoder Architecture}
\label{sec:cascaded_decoder}

Our framework features a cascaded decoder that progressively refines features across three sequential task-specific blocks, indexed by $k \in \{1, 2, 3\}$. These blocks correspond to the reconstruction tasks $t \in \{S, I, R\}$ (Semantic, Instance, RGB), respectively, aligning with our top-down masking strategy. As depicted in Figure~\ref{fig:framework}, the input to the decoder is the sequence of encoded visible tokens $\mathbf{H}_{enc} \in \mathbb{R}^{N_{vis} \times D}$.
Each decoder block $k$ is a standard Transformer decoder block, composed of self-attention, cross-attention, and feed-forward network layers. We use simple linear layers for the final prediction. For a given task $t$, the inputs to the decoder block $k$ are defined as follows:

\noindent\textbf{Query ($\mathbf{Q}^k$):} The encoded visible tokens $\mathbf{H}_{enc}$ are first combined with learnable mask tokens at the corresponding masked positions to reconstruct the full token sequence $\mathbf{H} \in \mathbb{R}^{3N\times D}$. This is analogous to the mask token insertion step in MAE~\cite{he2022masked}. Since the tokens for different tasks maintain fixed positions within this sequence, the query specific to task $t$ is obtained by slicing $\mathbf{H}$ along the sequence dimension based on pre-defined start and end indices $(i_t, j_t)$ as $\mathbf{Q}^k = \mathbf{H}[i_t:j_t]  \in \mathbb{R}^{N\times D}$.

\noindent\textbf{Key ($\mathbf{K}^k$) and Value ($\mathbf{V}^k$):} The key and value are formed by fusing the full token sequence $\mathbf{H}$ with the output features from the preceding block. Specifically, the output $\mathbf{F}^{k-1} \in \mathbb{R}^{N \times D}$ from block $k{-}1$ (initialized as $\mathbf{F}^0 = \mathbf{0}$) is placed back into its corresponding positions within $\mathbf{H}$ via element-wise addition:
\begin{equation}
  \mathbf{K}^k = \mathbf{V}^k = \mathbf{H} \oplus \text{scatter}(\mathbf{F}^{k-1}, i_{t_{k-1}}, j_{t_{k-1}}),
\end{equation}
where $\text{scatter}(\cdot)$ maps the $N$-dimensional output back to the corresponding task positions within the $3N$-dimensional sequence, and all other positions remain unchanged. This mechanism allows subsequent decoder blocks to attend to the refined features from earlier stages, establishing the cascaded information flow. 

The output feature of each task block, $\mathbf{F}^k \in \mathbb{R}^{N \times D}$, is obtained by applying the decoder block:
\begin{equation}
  \mathbf{F}^k = \text{DecoderBlock}^k ( \mathbf{Q}^k, \mathbf{K}^k, \mathbf{V}^k ).
\end{equation} 
Finally, a task-specific linear predictor generates the reconstruction for task $t$:
\begin{equation}
  \mathbf{\hat{I}}^t = \text{Predictor}^t(\mathbf{F}^k),
\end{equation}
where $\mathbf{\hat{I}}^t$ is the reconstructed output for the corresponding task. 
In contrast to parallel architectures, this progressive refinement allows each stage to build upon the last, enforcing a coarse-to-fine information flow that is crucial for learning hierarchical representations.

\subsection{Top-down Progressive Masking Strategy}
\label{sec:masking_strategy}

Our top-down progressive masking strategy facilitates hierarchical visual understanding through three pre-training phases using the following generated masks:
\begin{itemize}[leftmargin=2em]
    \item Semantic-guided mask ($\mathbf{M}_S$): Applies random masking within each semantic region, with the number of masked patches allocated to each region being proportional to its relative area.
    \item Instance-guided mask ($\mathbf{M}_I$): Guides masking based on instance information, prioritizing the occlusion of object regions over the background.
    \item Random mask ($\mathbf{M}_R$): Applies standard uniform random masking across the image with no structural guidance.
\end{itemize}

Below, we detail each phase of our masking strategy in their chronological training order, from the initial coarse-grained guidance to the final fine-grained refinement, and explain how they collectively contribute to hierarchical visual understanding.

For all three phases, the masking ratio for each granularity $m \in \{rgb, ins, sem\}$ is denoted as $r_m$, sampled from a Dirichlet distribution following MultiMAE~\cite{bachmann2022multimae}, such that $\sum_m r_m = r$, where $r$ is the overall masking ratio. This ensures a variable distribution of visible tokens across granularities while maintaining a fixed total masking budget. The three phases differ only in \textit{how} the $\lfloor r_m N \rfloor$ masked positions are selected for each granularity.

\noindent\textbf{Semantic-guided Masking for Scene-level Perception.}
  In this phase, we introduce semantic-guided masking based on the semantic regions. Given semantic masks $\mathbf{I}_{sem}$ with $C$ classes, we assign different masking weights to different semantic regions. Let $\Omega_c$ denote the set of patches belonging to semantic class $c$, and $w_c$ represent the corresponding class weight. The masking process can be formulated as:
  \begin{equation}
    \mathbf{M}_S^m = f_{sem}(\mathbf{I}_{sem}, r_m, \boldsymbol{w}) \in \{0,1\}^N
  \end{equation}
  where $f_{sem}$ is the semantic-guided masking function, and $\boldsymbol{w}$ are the class-specific weights. Specifically, $f_{sem}$ includes: 1) grouping patches according to their semantic labels to regions; 2) calculating the number of masked patches for each semantic region based on both region importance weights and region sizes; 3) randomly selecting patches within each region. For semantic class $c$, the number of masked patches is determined by:
  \begin{equation}
    |\mathbf{M}_S^m \cap \Omega_c| = \left\lfloor r_m N \cdot \frac{|\Omega_c|}{\sum_{k=1}^C |\Omega_k|} \right\rfloor.
  \end{equation}
  This region-proportional allocation ensures that the masking distribution reflects the spatial composition of the scene, preventing small but semantically important regions from being overlooked. Note that this formulation naturally extends to class-weighted sampling by replacing $|\Omega_c|$ with $w_c|\Omega_c|$, though we find equal weights sufficient in practice.

  \noindent\textbf{Instance-guided Masking for Object-level Understanding.}
  In this phase, we transition to instance-guided masking to promote object-centric learning. Given the instance masks $\mathbf{I}_{ins}$, we distribute the masked patches with emphasis on object regions. Let $\Omega_{obj}$ and $\Omega_{bg}$ denote the sets of patches belonging to object regions and background regions, respectively. The masking process can be formulated as:
  \begin{equation}
    \mathbf{M}_I^m = f_{ins}(\mathbf{I}_{ins}, r_m, \alpha) \in \{0,1\}^N,
  \end{equation}
  where $f_{ins}$ is our instance-guided masking function, and $\alpha$ controls the distribution of masks between object and background regions. Specifically, $f_{ins}$ includes: 1) identifying complete object instances from $\mathbf{I}_{ins}$; 2) selecting all or a subset of instances based on their size and spatial significance; 3) generating masks by randomly and separately selecting masked patches in object and background regions according to the ratio $\alpha$, which means
  \begin{align}
      &|\mathbf{M}_I^m \cap \Omega_{obj}| = \lfloor \alpha r_m N \rfloor,\\ \nonumber
      &|\mathbf{M}_I^m \cap \Omega_{bg}| = r_m N - \lfloor \alpha r_m N \rfloor.
  \end{align}
  We assign $\alpha > 0.5$ to prioritize masking object regions, ensuring that a larger portion of masked patches is allocated to object regions while maintaining some masking in the background for contextual learning.

  \noindent\textbf{Random Masking for Local Feature Understanding.}
  In the final phase, we employ standard random masking to refine fine-grained local feature learning. For each granularity $m \in \{rgb, ins, sem\}$, the masking ratio $r_m$ is sampled from a Dirichlet distribution following MultiMAE~\cite{bachmann2022multimae}, with a total masking ratio of $r$. The binary mask is then generated by uniformly sampling $\lfloor r_m N \rfloor$ positions:
  \begin{equation}
      \mathbf{M}_R^m = f_{rand}(r_m) \in \{0,1\}^N.
  \end{equation}
  Without any structural guidance, this phase compels the model to reconstruct arbitrary local regions, thereby strengthening its capacity for fine-grained detail understanding.

  \noindent\textbf{Progressive Training Schedule.}
  To facilitate seamless transitions across three pre-training phases, we propose a progressive training schedule that enables hierarchical knowledge accumulation while maintaining training stability. The final binary mask for each granularity $m$ is derived in two steps.

  First, we compute a continuous score map $\mathbf{S}^m \in [0,1]^N$ by linearly blending the three binary masks:
  \begin{equation}
  \mathbf{S}^m = (1-\alpha_I-\alpha_S)\mathbf{M}_R^m + \alpha_I \mathbf{M}_I^m + \alpha_S \mathbf{M}_S^m,
  \end{equation}
  where the coefficients satisfy $0 \leq \alpha_I, \alpha_S$ and $\alpha_I + \alpha_S \leq 1$.

  Then, the final binary mask $\mathbf{M}^m \in \{0,1\}^N$ is obtained by selecting the top $\lfloor r_m N \rfloor$ positions with the highest scores in $\mathbf{S}^m$:
  \begin{equation}
  \mathbf{M}^m = \text{top\text{-}k}(\mathbf{S}^m,\ \lfloor r_m N \rfloor),
  \end{equation}
  where $\text{top\text{-}k}(\cdot)$ returns a binary mask with 1 at the $\lfloor r_m N \rfloor$ highest-scoring positions and 0 elsewhere.

\begin{figure}[t]
    \centering
    \includegraphics[width=0.95\linewidth]{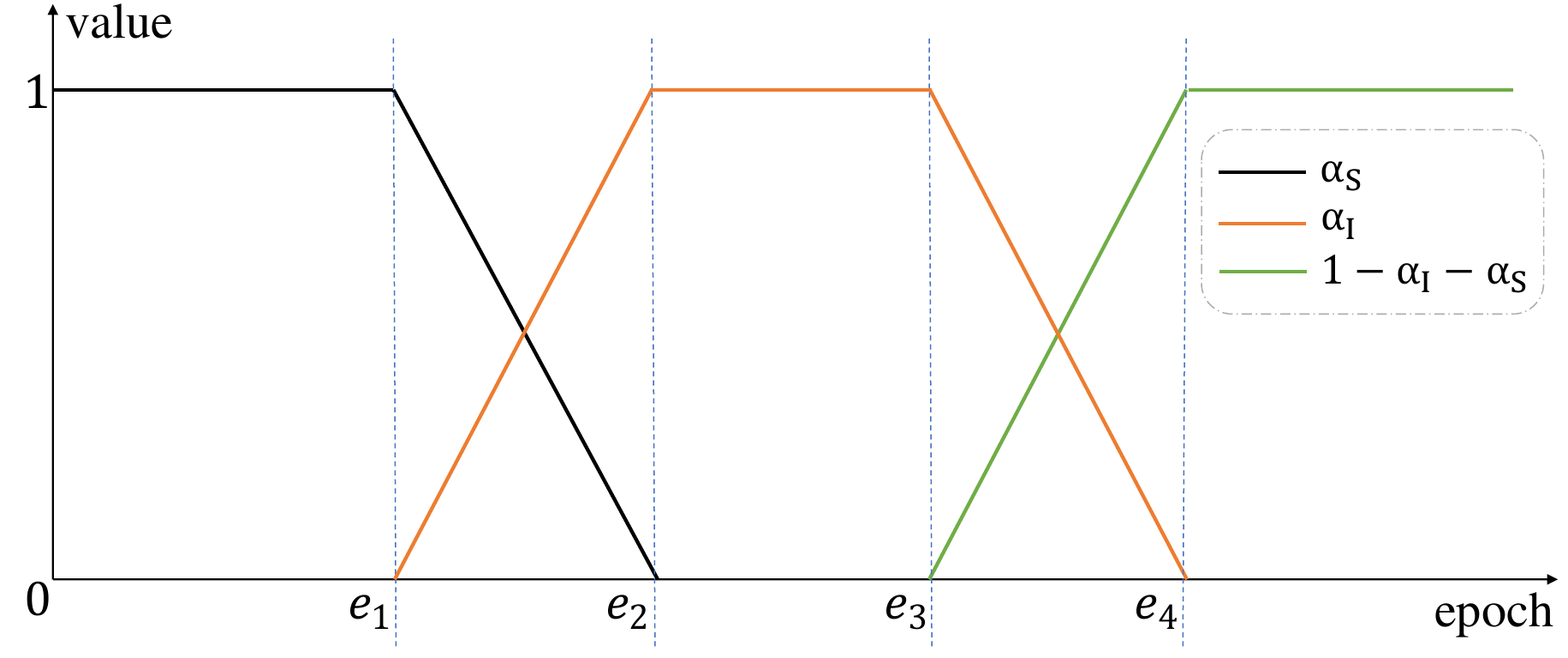}
     \caption{The variation curves of $\alpha_I$ and $\alpha_S$ during the training process.}
     \label{fig:transition}
  \end{figure}

  The coefficients $\alpha_I$ and $\alpha_S$ are dynamically adjusted during pre-training to transition the masking focus from semantic-guided to instance-guided smoothly and finally to random masking, as illustrated in Figure~\ref{fig:transition}. This progressive schedule creates a curriculum that guides the model to construct increasingly sophisticated visual representations, moving from coarse scene-level semantics to object-level structures, and finally to fine-grained local details.

\subsection{Training Objectives and Loss Functions}
\label{sec:loss}
Our framework is optimized with a multi-task objective to learn a comprehensive representation across distinct granularity levels. The overall loss combines three reconstruction losses, each aligned with a stage in our cascaded decoder:
\begin{equation}
\mathcal{L}_{\text{total}} = \lambda_{S}\mathcal{L}_{\text{S}} + \lambda_{I}\mathcal{L}_{\text{I}} + \lambda_{R}\mathcal{L}_{\text{R}},
\end{equation}
where $\mathcal{L}_{\text{S}}$ and $\mathcal{L}_{\text{I}}$ are the cross-entropy losses for semantic and instance mask prediction, respectively, while $\mathcal{L}_{\text{R}}$ is the mean squared error for RGB image reconstruction.
The coefficients $\lambda_{\text{S}}$, $\lambda_{\text{I}}$, and $\lambda_{\text{R}}$ are weighting factors that balance the contribution of each task, guiding the model to learn a coarse-to-fine hierarchical visual representation.

\begin{table*}[tb]
  \centering
  \caption{
  Performance comparison of self-supervised methods on ImageNet-1K. We report fine-tuning Top-1 accuracy (\%) of ViT-B. The input size is $224\times224$. PT Cost is the relative time to MAE (400 epochs), which is taken as 1.0. * indicates results reproduced using the official code. $\dagger$ means using multi-granular data as input for fine-tuning.
  }
  \label{tab:imagenet}
  \tablestyle{15pt}{1.2}
  \begin{tabular}{@{}lcccccc@{}}
    \toprule
    Method & Model & Modality & Masking & PT Epoch & PT Cost & Acc.\\
    \midrule
    Scratch  & ViT-B & $-$ & $-$&$-$&$-$&82.3\\
    \midrule
    MoCo v3~\cite{chenempirical} & ViT-B & RGB & $-$&300& $-$&83.2\\
    DINO~\cite{caron2021emerging} & ViT-B & RGB& $-$ & 300 &$-$&82.8\\
    BEiT~\cite{bao2021beit} & ViT-B & RGB &Random&800& $\sim$7.0$\times$ &83.2\\
    MAE~\cite{he2022masked}*  & ViT-B & RGB&Random&400& $\sim$1.0$\times$ &82.9\\
    MAE~\cite{he2022masked}  & ViT-B & RGB&Random&1600& $\sim$4.0$\times$ &83.6\\
    iBOT~\cite{zhou2022image} & ViT-B & RGB &Random&1600& $\sim$5.7$\times$ &84.0\\
    UnMAE\cite{li2022uniform} & ViT-B & RGB&Uniform&200& $-$&82.9\\
    CAE~\cite{chen2024context} & ViT-B & RGB&Random&800& $\sim$4.6$\times$ &83.6\\
    MaskFeat~\cite{cheng2022masked} & ViT-B & RGB&Random&1600& $\sim$20.1$\times$ &84.0\\
    SemMAE~\cite{li2022semmae} & ViT-B & RGB&Semantic&800& $-$ &83.3\\
    AutoMAE~\cite{chen2023improving} & ViT-B & RGB&Semantic&800& $-$&83.3\\
    ConMIM~\cite{yi2023masked} & ViT-B & RGB&Random&800& $\sim$4.4$\times$ &83.7 \\
    MIRL~\cite{NEURIPS2023_b3bac97f} & ViT-B & RGB&Random&800& $-$ &84.1 \\
    ROPIM~\cite{haghighat2024pretraining} & ViT-B & RGB&Random&800& $\sim$10.4$\times$ &84.0 \\
    MFM~\cite{xie2022masked}  & ViT-B & RGB/Frequency&Random&300& $\sim$1.1$\times$ &83.1\\
    MultiMAE*~\cite{bachmann2022multimae} & ViT-B & RGB/Dep./Sem. &Random&400& $\sim$1.3$\times$ &82.7\\
    MultiMAE~\cite{bachmann2022multimae} & ViT-B & RGB/Dep./Sem. &Random&1600& $\sim$5.2$\times$ &83.3\\
    C2FMAE & ViT-B & RGB/Inst./Sem. &Progressive&400& $\sim$1.3$\times$ &83.7\\
    C2FMAE & ViT-B & RGB/Inst./Sem. &Progressive&1600& $\sim$5.2$\times$ &\textbf{84.2}\\
    \midrule
    C2FMAE$\dagger$ & ViT-B & RGB/Inst./Sem. &Progressive&1600& $\sim$5.2$\times$ &\textbf{84.4}\\
  \bottomrule
  \end{tabular}
\end{table*}

\section{Experiments}
\label{sec:experiments}

This section outlines our experimental validation. First, we show the detailed experimental settings for the pre-training and downstream task evaluations. Then, we present the main experimental results, benchmarking C2FMAE against state-of-the-art methods and evaluating its robustness. Following this, we conduct comprehensive ablation studies to dissect the contributions of key components within our framework. Finally, qualitative visualizations are provided to intuitively demonstrate the efficacy of our proposed approach. All experiments were conducted on a server with $8\times$ NVIDIA Tesla-A100 GPUs.

\subsection{Experiment Settings}
  
  \label{sec:d}
  
  \noindent \textbf{Pretraining.}
  Our framework employs Vision Transformer~\cite{dosovitskiy2021an} as the backbone network, processing $224 \times 224$ input images from ImageNet-1K~\cite{russakovsky2015imagenet}. We utilize the ViT-B model with patch size of 16, adopting a masking ratio of 1/6. The input data include three modalities: RGB images, instance masks, and semantic masks. We design a dedicated decoder comprising one cross-attention and two self-attention transformer blocks, featuring a dimensionality of 256 with 8 attention heads.
  The models are trained for 400 or 1600 epochs, including a 40-epoch warmup phase, with a total batch size of 2048. We employ the AdamW~\cite{loshchilov2018decoupled} optimizer with a base learning rate of $1 \times 10^{-4}$, linearly scaled as $\text{lr = base\_lr} \times \text{batch\_size} / 256$. The optimization parameters include a weight decay of 0.05, momentum parameters $\beta_1 = 0.9$ and $\beta_2 = 0.95$, and a cosine learning rate decay schedule. Standard data augmentations such as random cropping and horizontal flipping are applied. All Transformer blocks are initialized using Xavier uniform initialization~\cite{glorot2010understanding}, following the MAE~\cite{he2022masked} approach.
  
  \noindent \textbf{Image Classification.}
  We conduct end-to-end supervised fine-tuning on the ImageNet-1K dataset at $224 \times 224$ resolution, adhering to standard practices for fair method comparison. For ViT-B, we train for 100 epochs with 5 warmup epochs, employing base learning rates of 1$e-$3/5$e-$4 and layer-wise learning rate decay of 0.7/0.65 for 400/1600 epochs, respectively.
  The training configuration maintains a batch size of 2048 and a drop path rate of 0.1~\cite{huang2016deep}. Robust data augmentation techniques are applied, including label smoothing~\cite{szegedy2016rethinking}, mixup~\cite{zhang2017mixup}, cutmix~\cite{yun2019cutmix}, and randAugment~\cite{cubuk2020randaugment}. Following MAE, we replace class tokens with global pooling features during fine-tuning. The learning rate adheres to the linear scaling rule: $\text{lr = base\_lr} \times \text{batch\_size} / 256$.

  \begin{table}[tb]
    \centering
    \caption{
    Object detection and instance segmentation results on the COCO dataset, with 
  evaluation metrics of AP$^{b}$ ($\%$) and AP$^{m}$ ($\%$). The pre-trained ViT-B backbone is integrated into the Mask R-CNN framework for end-to-end fine-tuning.
    }
    \label{tab:coco}
  \tablestyle{13pt}{1.2}
    \begin{tabular}{@{}lcccc@{}}
      \toprule
      Method & Model &PT Epoch  & AP$^{b}$ & AP$^{m}$\\
      \hline
       MoCo v3& ViT-B & 300&47.3& 42.2 \\
       BEiT& ViT-B & 800 & 35.6& 32.6\\
       MAE& ViT-B & 1600 & 48.3& 42.5\\
       CAE& ViT-B & 800 & 49.8& 43.9\\
       iBOT& ViT-B & 1600 & 48.3& 42.7 \\
       ConMIM& ViT-B & 800 & 47.8&42.5 \\
       MIRL& ViT-B & 800 & 49.3&43.7 \\
       MultiMAE& ViT-B & 1600& 48.1& 42.2\\
       C2FMAE& ViT-B & 1600& \textbf{50.1} & \textbf{44.1} \\
    \bottomrule
    \end{tabular}
  \end{table}
  
  \noindent \textbf{Object Detection and Instance Segmentation.}
  We integrate the pre-trained ViT backbone into the Mask R-CNN~\cite{he2017mask} framework, conducting fine-tuning on the COCO~\cite{lin2014microsoft} dataset using the MMDetection~\cite{chen2019mmdetection} implementation. The adaptation involves multi-scale training, resizing images to have a short side between 480 and 800 and a long side no greater than 1333. We employ the AdamW optimizer with a learning rate of 3$e-$3, weight decay of 0.05, and total batch size of 16. Layer-wise decay rates are 0.75, with drop path rates of 0.2, respectively. We utilize a 1$\times$ training schedule of 12 epochs, decaying the learning rate by 10$\times$ at epochs 9 and 11. Performance is evaluated on COCO val2017 using bounding box AP$^{b}$ and mask AP$^{m}$ metrics.
  
  \noindent \textbf{Semantic Segmentation.}
  We incorporate the pre-trained ViT-B backbone into the UperNet~\cite{xiao2018unified} architecture for semantic segmentation on the ADE20K~\cite{zhou2017scene} dataset. The fine-tuning process spans 160k iterations with $512 \times 512$ input resolution, utilizing the AdamW optimizer.
  Key training parameters include a base learning rate of 4$e-$4, weight decay of 0.05, and batch size of 16. The learning rate follows a warmup of 1500 iterations before linear decay. Segmentation performance is evaluated using mIoU on the validation set.

  \begin{table}[tb]
    \centering
    \caption{
    Semantic segmentation results on the ADE20K dataset, with 
  evaluation metrics of mIoU. The pre-trained ViT-B backbone is integrated with  UperNet for end-to-end fine-tuning.
    }
    \label{tab:ade20k}
  \tablestyle{13pt}{1.2}
    \begin{tabular}{@{}lccc@{}}
      \toprule
      Method& Model & PT Epoch  & mIoU\\
      \hline
       MoCo v3& ViT-B & 300 & 47.3\\
       BEiT& ViT-B & 800  & 47.1\\
       MAE& ViT-B &1600  & 48.1\\
       MaskFeat& ViT-B &1600 &48.8 \\
       CAE& ViT-B & 800 & 48.8 \\
       SemMAE& ViT-B & 800 & 46.3 \\
       ConMIM& ViT-B &1600 & 46.0 \\
       ROPIM & ViT-B & 300 & 48.5 \\
       MultiMAE& ViT-B & 1600  & 47.8 \\
       C2FMAE& ViT-B & 1600 & \textbf{49.1} \\
    \bottomrule
    \end{tabular}
  \end{table}

\subsection{Main Results}

\noindent \textbf{Image Classification.}
We first evaluate our method on ImageNet-1K, comparing its fine-tuning top-1 accuracy against key baselines and state-of-the-art methods. 
Alongside the foundational MAE baseline, MultiMAE serves as a particularly strong counterpart, as it also leverages multi-modal data but employs a parallel decoder and a simple random masking strategy.

As shown in Table~\ref{tab:imagenet}, our C2FMAE achieves fine-tuning accuracies of 83.7\% and 84.2\% after 400 and 1600 pre-training epochs, respectively. This represents a significant improvement over both the baseline MAE (+0.8\%/+0.6\%) and MultiMAE (+1.0\%/+0.9\%). Notably, MultiMAE underperforms the simpler MAE, suggesting that its parallel processing of modalities fails to effectively integrate hierarchical information. In contrast, C2FMAE's superior performance validates that our coarse-to-fine framework, with its cascaded decoder and progressive masking, successfully learns more powerful hierarchical representations. Furthermore, we explored fine-tuning on ImageNet-1K using multi-granular data as input, achieving a final accuracy of 84.4\%, which is competitive with current state-of-the-art methods. This indicates that within our framework, the different data granularities can also synergistically boost downstream task performance.

In terms of computational cost, C2FMAE's training time is nearly identical to MultiMAE's and only about 1.3 times that of MAE, primarily due to the increased number of input tokens. The overhead from the cascaded decoder and progressive mask generation is negligible. Crucially, our 400-epoch model already surpasses the performance of MAE's 1600-epoch model (83.7\% vs. 83.6\%). This demonstrates that C2FMAE not only achieves higher accuracy but does so more efficiently, constructing rich hierarchical representations in a fraction of the training time.

\noindent \textbf{Object Detection and Instance Segmentation.} 
To evaluate the transferability of our approach to dense downstream tasks, we conduct experiments on object detection and instance segmentation using the COCO dataset. As shown in Table~\ref{tab:coco}, our pre-trained model achieves significant gains over key baselines. Specifically, C2FMAE outperforms MAE by +1.8 AP$^b$ and +1.6 AP$^m$, and surpasses MultiMAE by +2.0 AP$^b$ and +1.9 AP$^m$ in object detection and instance segmentation, respectively. These substantial improvements underscore that our coarse-to-fine pre-training framework effectively enhances the model's ability to capture the hierarchical features crucial for complex dense prediction tasks.

\noindent \textbf{Semantic Segmentation.}
We further extend our evaluation to semantic segmentation on the challenging ADE20K benchmark. As reported in Table~\ref{tab:ade20k}, C2FMAE achieves a superior mIoU of 49.1\%, significantly outperforming the baseline MAE by +1.0\% and the multi-modal MultiMAE by +1.3\%. This result offers critical insight into the efficacy of our method: unlike MultiMAE, which treats semantic information as an isolated modality in a parallel decoder, C2FMAE's cascaded structure explicitly enforces a dependency where high-level semantics guide the refinement of low-level features. This aligns perfectly with the intrinsic demands of semantic segmentation, which requires both global context for accurate classification and fine-grained spatial encodings for precise boundary delineation. 

\begin{table}[tb]
  \centering
  \caption{Robustness results of ViT-B on various ImageNet OOD variants. Top-1 accuracy is reported for all datasets, except for ImageNet-C, where the mean corruption error (mCE) is the evaluation metric. And (1-mCE) is used to calculate the average score. 
  }
  \label{tab:robustness}
  % \small
  \tablestyle{8pt}{1.2}
  \begin{tabular}{@{}lccccc@{}}
    \toprule
    Method & IN-A & IN-R &IN-S&IN-C $\downarrow$& Score \\
    \midrule
     MAE   & \textbf{35.9} & 48.3 & 34.5 & 51.7 &41.8\\
     MFM  & 32.7 & 48.6 & 34.8 & \textbf{47.5}& 42.2 \\
     MultiMAE & 33.2 & 49.9 & 37.2 & 49.6 & 42.7 \\
     C2FMAE  & 35.2 & \textbf{50.6} & \textbf{37.4} & 48.8 & \textbf{43.6} \\
  \bottomrule
  \end{tabular}
\end{table}

\begin{table}[tb]
  \centering
  \caption{Component analysis of C2FMAE on ImageNet-1K. Starting from a MultiMAE baseline, we incrementally add our proposed components. 
  }
  \label{tab:component_analysis}
  \tablestyle{5pt}{1.2}
  \begin{tabular}{y{150}x{40}}
  \toprule
  Configuration & Top-1 \\
  \midrule
  Baseline (MultiMAE) & 82.7 \\
  + Our Dataset (R+I+S) & 83.0(+0.3) \\
  + Cascaded Decoder & 83.3(+0.3) \\
  + Progressive Masking (Ours) & \textbf{83.7}(+0.4) \\
  \bottomrule
  \end{tabular}
\end{table}

\noindent \textbf{Robustness Evaluation.}
We assess the robustness of our methods across various out-of-distribution (OOD) ImageNet benchmarks, including ImageNet-A~\cite{hendrycks2021natural}, ImageNet-R~\cite{hendrycks2021many}, ImageNet-Sketch~\cite{wang2019learning}, and ImageNet-C~\cite{hendrycks2018benchmarking}. Top-1 accuracy is the primary evaluation metric for all datasets, except for ImageNet-C, where we report mean corruption error (mCE). To derive the final robustness score, we calculate \(1 - \text{mCE}\) and take the average across all tested datasets.
As shown in Table~\ref{tab:robustness}, our method demonstrates superior robustness compared to other approaches. We achieve improvements across all four OOD datasets, with the most significant gains observed on ImageNet-R and ImageNet-Sketch. Specifically, our model shows an average score increase of 1.8\% and 0.9\% over MAE and MultiMAE, highlighting that our framework helps the model learn more robust visual representations, significantly enhancing its robustness against OOD data.

\subsection{Ablation studies}
In this section, we perform ablation studies to evaluate the key components of our framework. All models are pre-trained for 400 epochs using ViT-B.
We first conduct a step-by-step component analysis. Starting from the MultiMAE baseline, we incrementally add our proposed components. The results are presented in Table~\ref{tab:component_analysis}. This incremental performance gain at each step strongly verifies the effectiveness and synergistic nature of our proposed components. 
Then, we further investigate the hyper-parameters and design choices for different components, including input tokens, masking strategy, decoder design, input modality, and loss weight. The results are summarized in Table~\ref{tab:ablations}.
All models are pre-trained for 400 epochs using ViT-B, and we report fine-tuning Top-1 accuracy on ImageNet-1K validation set.

\noindent \textbf{Input Tokens.} 
We use the number of input tokens as a proxy for the mask ratio. With a total of $196\times3$ patches, input token counts of 98, 147, and 196 correspond to mask ratios of 0.833, 0.75, and 0.667, respectively. As shown in Table~\ref{tab:inputtokens}, our framework achieves optimal performance at the highest mask ratio (fewest input tokens). This aligns with the core principle of masked image modeling, where a more challenging reconstruction task from limited visible context compels the model to learn more holistic and generalizable representations. This approach yields a dual benefit: it not only improves downstream performance but also enhances training efficiency by reducing the computational load on the encoder.

  \begin{table*}[tb]
  \centering
  
  \caption{{C2FMAE ablation experiments} on ImageNet-1K. The fine-tuning Top-1 accuracy (\%) is reported. 
  The default settings include 98 for the number of input tokens, progressive masking with the order of SG→IG→RD, a cascaded decoder with cross-attention and task sequence of S→I→R, R+I+S for the input modality, and $\lambda_S=\lambda_I=\lambda_R=1$ for loss weight.
  The selected settings are \underline{underlined}.}
  \label{tab:ablations} 
  
  %#################################################
  % Row 1
  %#################################################
  \subfloat[{Input Tokens}\label{tab:inputtokens}]{
      \begin{minipage}[t]{0.32\linewidth} % 使用 [t] 确保顶部对齐
      \centering
      \tablestyle{10pt}{1.2}
      \begin{tabular}{x{34}x{22}}
      \toprule
      Number & Top-1 \\
      \midrule
      98 & \textbf{\underline{83.7}} \\
      147 & 83.7 \\
      196 & 83.5 \\
      \bottomrule
      \end{tabular}
      \end{minipage}
  }
  \hfill
  \subfloat[{Single Masking strategy}\label{tab:masking_stra}]{
      \begin{minipage}[t]{0.32\linewidth}
      \centering
      \tablestyle{10pt}{1.2}
      \begin{tabular}{x{34}x{22}}
      \toprule
      Type & Top-1  \\
      \midrule
      RD & 83.3  \\
      IG & \textbf{83.5}  \\
      SG & 83.4  \\
      % SG→IG→RD & \textbf{83.7}  \\
      \bottomrule
      \end{tabular}
      \end{minipage}
  }
  \hfill
  \subfloat[{Masking order}\label{tab:mask_stragety}]{
      \begin{minipage}[t]{0.32\linewidth}
      \centering
      \tablestyle{8pt}{1.2}
      \begin{tabular}{x{40}x{22}}
      \toprule
      Type & Top-1 \\
      \midrule
      IG→SG→RD & 83.5 \\
      RD→IG→SG & 83.5 \\
      SG→IG→RD & \textbf{\underline{83.7}} \\
      \bottomrule
      \end{tabular}
      \end{minipage}
  }
  
  %#################################################
  % Row 2 - Add vertical space
  %#################################################
  \vspace{1em} 
  
  \subfloat[{Decoder Design}\label{tab:stage_trans}]{
      \begin{minipage}[t]{0.32\linewidth}
      \centering
      \tablestyle{10pt}{1.2}
      \begin{tabular}{x{34}x{22}}
      \toprule
      Type & Top-1  \\
      \midrule
      Parallel & 83.3 \\
      \hline
      w/o CA & 83.2 \\
      \hline
      R→I→S & 83.5 \\
      S→I→R & \textbf{\underline{83.7}}  \\
      \bottomrule
      \end{tabular}
      \end{minipage}
  }
  \hfill
  \subfloat[{Input modality}\label{tab:input_modality}]{
      \begin{minipage}[t]{0.32\linewidth}
      \centering
      \tablestyle{5pt}{1.2}
      \begin{tabular}{x{24}x{40}x{20}}
      \toprule
      Modality & Masking & Top-1 \\
      \midrule
      RGB & RD&82.9  \\
      R+S & RD&83.2  \\
      R+I & RD&83.1  \\
      R+I+S & RD&83.3  \\
      R+I+S & SG→IG→RD&\textbf{\underline{83.7}}  \\
      \bottomrule
      \end{tabular}
      \end{minipage}
  }
  \hfill
  \subfloat[{Loss weight}\label{tab:loss_weight}]{
      \begin{minipage}[t]{0.32\linewidth}
      \centering
      \tablestyle{10pt}{1.2}
      \begin{tabular}{x{34}x{22}}
      \toprule
      $\lambda_S, \lambda_I, \lambda_R$ & Top-1 \\
      \midrule
      1, 1, 1 & \textbf{\underline{83.7}}  \\
      1, 1, 2 & \textbf{83.7}  \\
      1, 2, 1 & 83.4  \\
      2, 1, 1 & 83.5  \\
      \bottomrule
      \end{tabular}
      \end{minipage}
  }
\end{table*}

\noindent \textbf{Mask strategy.} 
To verify the effectiveness of our proposed masking strategy, we first evaluate the three core masking approaches, Random Masking (RD), Instance-Guided Masking (IG), and Semantic-Guided Masking (SG) individually. The results in Table~\ref{tab:masking_stra} demonstrate that both IG and SG strategies contribute positively to representation learning compared to RD. Building on this, to assess the impact of dynamically sequencing these strategies, we test our progressive masking approach with different orders.
As evidenced in Table~\ref{tab:mask_stragety}, our progressive masking strategy demonstrates superior efficacy over static random masking in facilitating hierarchical visual representations. The top-down masking order (SG→IG→RG), well-aligned with our cascaded decoder architecture, yields further performance gains. Our progressive masking scheme enables the model to progressively construct visual features through semantic abstraction hierarchies, beginning with high-level semantic concepts and gradually refining localized detail representations.

\noindent \textbf{Decoder Design.} 
We evaluate the effectiveness of cascaded decoders within our framework, and further investigate the impacts of cross-attention mechanisms and task sequencing in reconstruction objectives. As demonstrated in Table~\ref{tab:stage_trans}, the cascaded decoder architecture significantly enhances the hierarchical construction of visual representations compared to the parallel decoder. The task sequence, starting with semantic mask prediction, followed by instance mask prediction, and ending with RGB image reconstruction, effectively supports hierarchical feature development. Notably, cross-attention plays a critical role in the decoder design by facilitating effective cross-modal interaction.

\begin{figure*}[t]
  \centering
   \includegraphics[width=\linewidth]{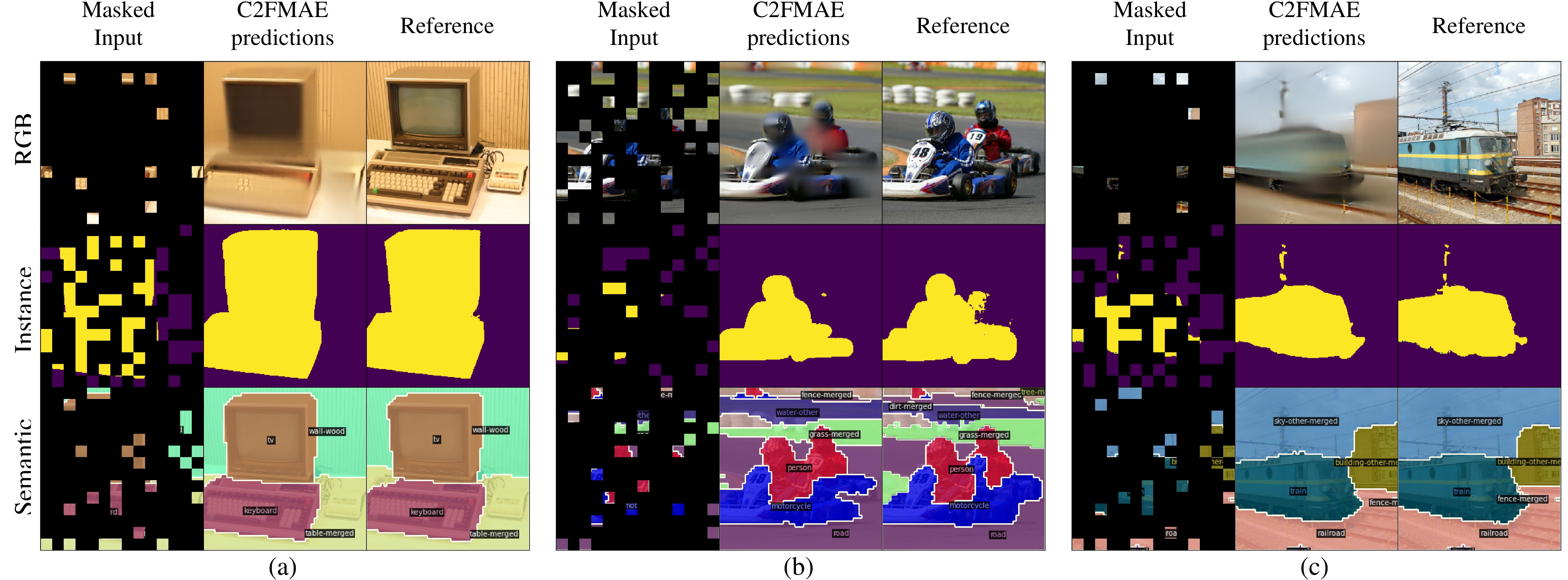}

   \caption{Predictions of C2FMAE on masked multi-granular data. All the tested images are from the ImageNet-1K validation set and masked with the random masking strategy. }
   \label{fig:vis1}
\end{figure*}

\begin{figure*}[t]
  \centering
   \includegraphics[width=\linewidth]{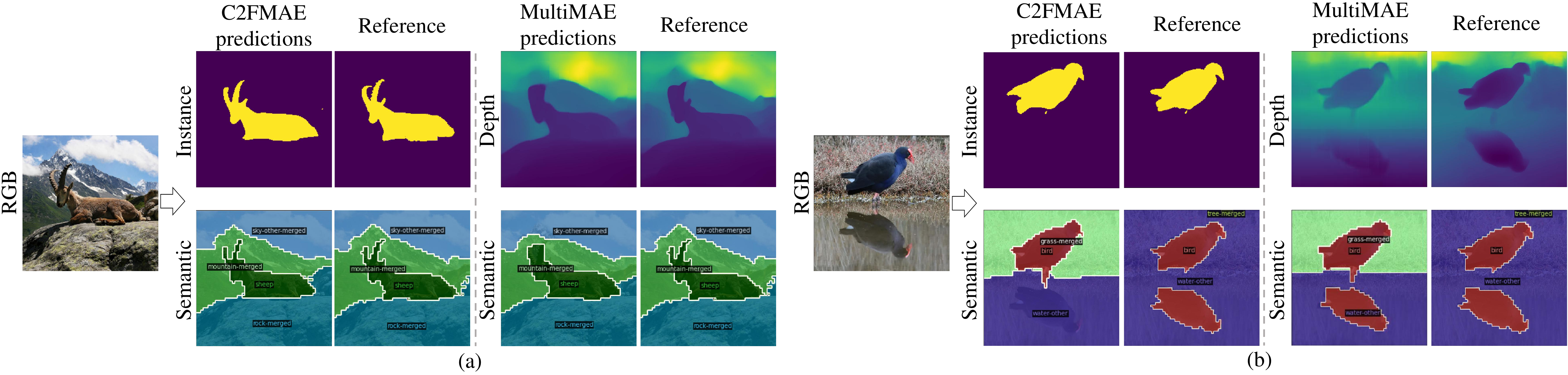}
  \caption{Single-modal prediction of C2FMAE and MultiMAE on the ImageNet-1K validation set.}
   \label{fig:vis2}
\end{figure*}

\begin{figure*}[t]
  \centering
   \includegraphics[width=\linewidth]{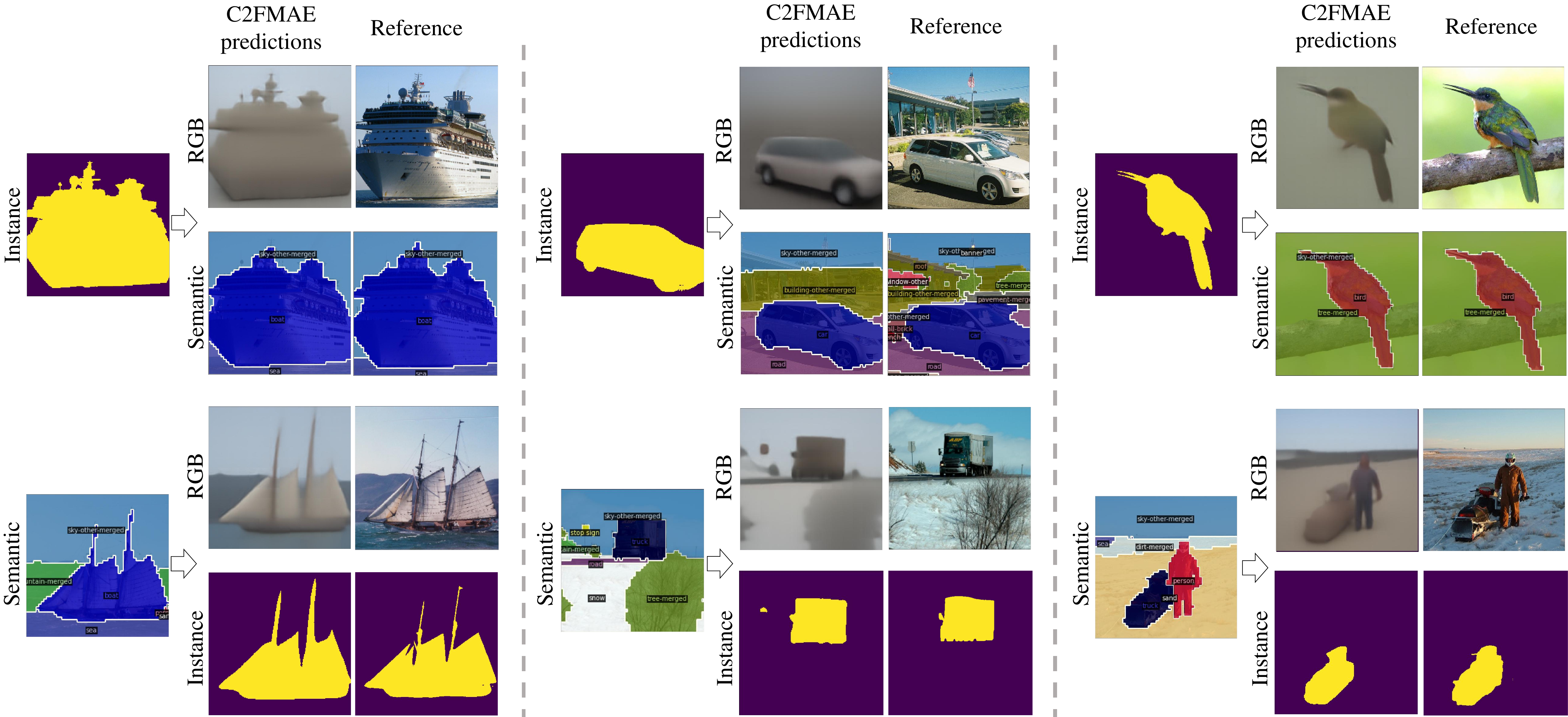}
  \caption{Single-modal prediction of C2FMAE with instance masks and semantic masks.}
   \label{fig:vis_single_seg}
\end{figure*}

\noindent \textbf{Input modality.} 
We conduct an ablation study to isolate the contribution of each data granularity. To ensure a fair comparison when certain granularities are omitted, we use a standard random masking strategy and our cascaded decoder for all configurations in this experiment, as our progressive masking strategy inherently requires the presence of all granularities. The results in Table~\ref{tab:input_modality} show that adding either semantic (R+S) or instance (R+I) masks improves performance over using RGB alone. Crucially, combining all three granularities (R+I+S) yields the best result, demonstrating a clear synergistic effect where different levels of abstraction collaboratively enhance the final representation. Furthermore, applying our progressive masking strategy (SG→IG→RD) to the full multi-granular input further boosts the accuracy to 83.7\%, proving that our carefully designed curriculum is essential to fully unleash the potential of the hierarchical data.

\noindent \textbf{Loss weight.} We investigate the impact of different weighting schemes for the semantic ($\lambda_S$), instance ($\lambda_I$), and RGB ($\lambda_R$) reconstruction losses. As shown in Table~\ref{tab:loss_weight}, we observe that when RGB reconstruction is the dominant task (e.g., $\lambda_R \geq $ $\lambda_S$, $\lambda_I$), downstream task performance is generally strong, likely because RGB images implicitly contain the richest information. For simplicity, we adopt equal weights ($\lambda_S=\lambda_I=\lambda_R=1$) as our default setting.

\subsection{Visualizations}

To highlight the improvements in representation learning achieved by our dataset and framework, we visualize the reconstruction results of our pre-trained auto-encoder across different data modalities. All images are from the ImageNet-1K validation set.
Figure~\ref{fig:vis1} illustrates the reconstruction capabilities of C2FMAE on masked multi-granular data. As shown, our model effectively reconstructs each granular data under different input conditions. Thanks to the interactivity and complementarity between the multi-granular dataset we have constructed, our model performs well in predicting fine details. Moreover, in some cases, it can even correct errors in the pseudo-labels, as seen in the instance prediction in Figure~\ref{fig:vis1}(b). 

We further explore using single-modal data to predict data from other modalities, \eg leveraging fine-grained RGB images to predict coarse-grained instance and semantic masks. The results shown in Figure~\ref{fig:vis2} highlight the transferability of our method across different modalities. Compared to MultiMAE, our model not only delivers superior performance in predicting finer details but also demonstrates greater robustness, \eg being unaffected by reflections of birds in water in Figure~\ref{fig:vis2}(b). 

Finally, we investigate the generative capability using sparse mask inputs to predict other modalities (Figure~\ref{fig:vis_single_seg}). This experiment reveals the distinct roles of different data granularities. When conditioned on \textit{Instance Masks}, which encode object shape but lack semantic content, the predicted RGB images accurately preserve object contours but resemble untextured 3D models, while the predicted semantic masks correctly infer object categories from structural cues. Conversely, when conditioned on \textit{Semantic Masks}, which provide categorical context, the predicted RGB images exhibit plausible textures corresponding to each semantic region, and the predicted instance masks accurately delineate the primary objects. This contrast highlights C2FMAE's ability to disentangle structural geometry (from instances) and semantic texture (from semantics), and to leverage their complementarity for comprehensive visual understanding.

\section{Conclusion}
In this paper, we propose C2FMAE, a coarse-to-fine vision pre-training framework that addresses the fundamental limitation of existing self-supervised methods, \ie, the inability to simultaneously capture hierarchical visual representations. By integrating a cascaded decoder and progressive masking strategy across RGB images, instance masks, and semantic masks, C2FMAE explicitly enforces a top-down information flow from scene-level semantics to pixel-level details. Extensive experiments demonstrate that C2FMAE achieves state-of-the-art performance on ImageNet-1K classification, COCO detection/segmentation, and ADE20K semantic segmentation, while maintaining training efficiency comparable to existing methods. The large-scale multi-granular dataset we constructed provides a valuable resource for future research in hierarchical visual understanding, as well as broader domains such as weakly-supervised dense prediction and controllable generative modeling.

\section*{Acknowledgments}
This work was supported in part by the National Natural Science Foundation of China under Grant 62402251 and Grant 62472238.

\bibliography{ref}
\bibliographystyle{IEEEtran}

\end{document}

\title{From Semantics to Pixels: Coarse-to-Fine Masked Autoencoders for Hierarchical Representation Learning}

\author{Wenzhao Xiang, Yue Wu, Hongyang Yu, Feng Gao, Fan Yang, Xilin Chen,~\IEEEmembership{Fellow,~IEEE}
        % <-this % stops a space
\thanks{Wenzhao Xiang (xiangwenzhao22@mails.ucas.ac.cn), Yue Wu (wuyue221@mails.ucas.ac.cn), and Xilin Chen (xlchen@ict.ac.cn) are with the Key Laboratory of Intelligent Information Processing, Institute of Computing Technology, Chinese Academy of Sciences, Beijing 100190, China, and also with the University of Chinese Academy of Sciences, Beijing 100049, China. Wenzhao Xiang, Yue Wu, and Hongyang Yu (yuhy01@pcl.ac.cn) are with Pengcheng Laboratory, Shenzhen 518108, China. Feng Gao (gaof@pku.edu.cn) and Fan Yang (fyang.eecs@pku.edu.cn) are with Peking University, School of Arts, No.5 Yiheyuan Road, Haidian District, Beijing 100871, China.}% <-this % stops a space
\thanks{Manuscript received April 19, 2021; revised August 16, 2021.}}

\markboth{Journal of \LaTeX\ Class Files,~Vol.~14, No.~8, August~2021}%
{Shell \MakeLowercase{\textit{et al.}}: A Sample Article Using IEEEtran.cls for IEEE Journals}


{

\clearpage
 \appendix
 \section*{Illustration of Our Dataset}
  \label{sec:a1}
  In Figure~\ref{fig:dataset}, we randomly display images representing various categories from our dataset. Each cell is structured in three columns: the first column presents the original RGB image, followed by its corresponding instance masks and semantic masks. The visualization effectively illustrates our dataset's exceptional segmentation precision across diverse classes of images.
  \begin{figure*}[htbp]
      \centering
      \includegraphics[width=0.95\textwidth]{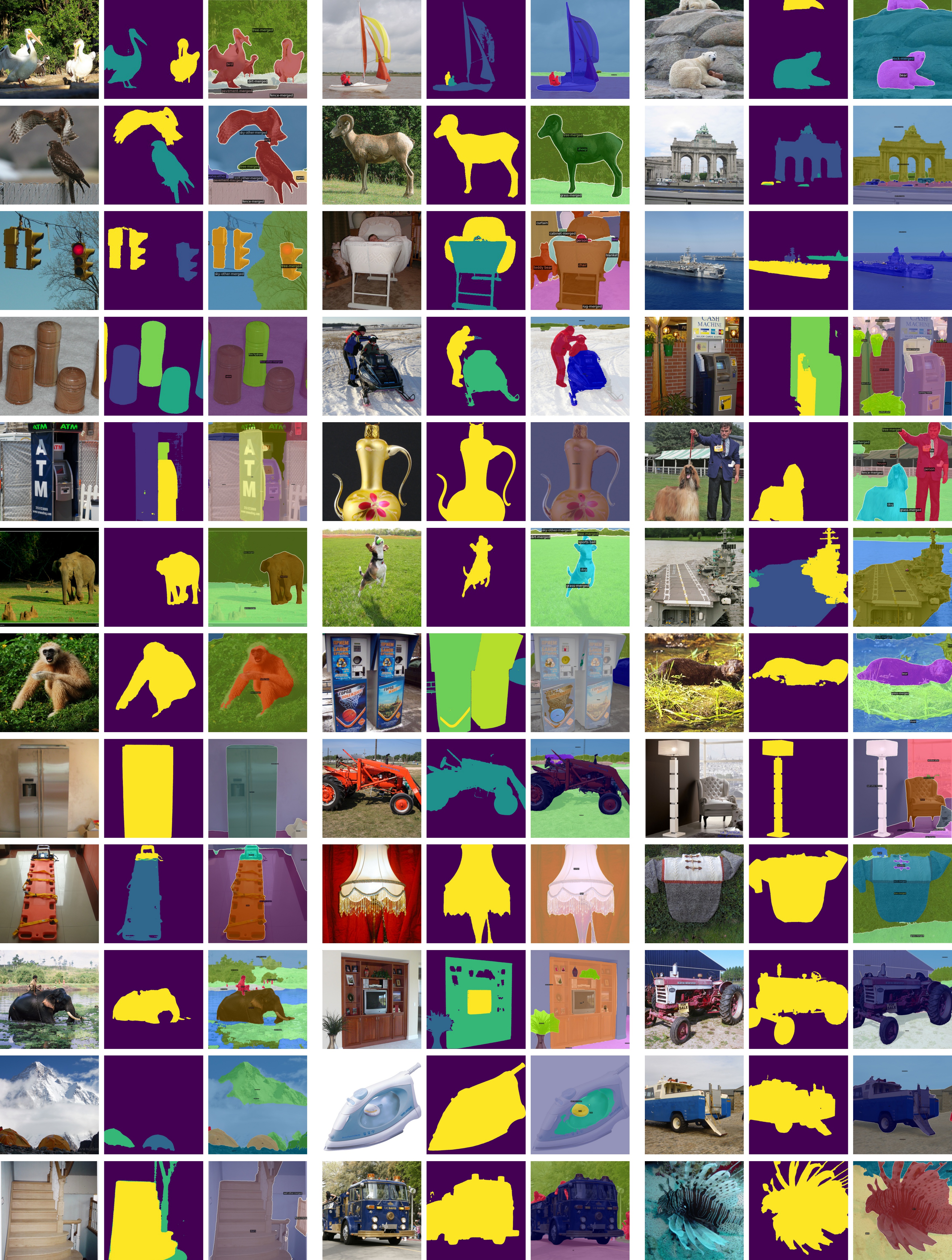}
      \caption{Illustration of the proposed Dataset.}
      \label{fig:dataset}
\end{figure*}

\section*{Detailed Experiment Settings}
  \label{sec:b}
  
  This appendix provides the complete hyperparameter configurations for all experiments.

  \noindent\textbf{Pretraining.}
  The full pre-training configuration is listed in Table~\ref{tab:impl_pretraining}. All models are pre-trained on 8$\times$ NVIDIA Tesla-A100 GPUs.
  
  \begin{table}[htp]
  \centering
  \caption{{Pre-training settings.}}
  \label{tab:impl_pretraining} 
  \begin{tabular}{y{95}|x{110}}
  \shline
  config & ViT-B \\
  \shline
  optimizer & AdamW \\
  base learning rate & 1e-4 \\
  weight decay & 0.05 \\
  optimizer momentum & $\beta_1, \beta_2{=}$ 0.9, 0.95 \\
  batch size & 2048 \\
  learning rate schedule & cosine decay \\
  pre-training epochs & 400/1600 \\
  warmup epochs & 10/40 \\
  augmentation & random cropping\& \ \ \ \  horizontal flip \\
  mask ratio &  1/6  \\
  pre-training resolution & 224 $\times$ 224 \\
  \shline
  \end{tabular}
  \end{table}
  
  \noindent\textbf{Image Classification.}
  The fine-tuning configuration for ImageNet-1K classification is detailed in Table~\ref{tab:impl_finetuning}. We follow the standard end-to-end fine-tuning protocol of MAE~\cite{he2022masked}.

  \begin{table}[h]
  \centering
  \caption{{Fine-tuning settings for image classification.}}
  \label{tab:impl_finetuning} 
  \begin{tabular}{y{106}|x{98}}
  \shline
  config & ViT-B \\
  \shline
  optimizer & AdamW \\
  base learning rate & 1e-3(400e);5e-4(1600e) \\
  weight decay & 0.05 \\
  optimizer momentum & $\beta_1, \beta_2{=}$ 0.9, 0.999 \\
  layer-wise decay & 0.7(400e);0.65(1600e) \\
  batch size & 2048 \\
  learning rate schedule & cosine decay \\
  training epochs & 100 \\
  warmup epochs & 5 \\
  augmentation & RandAug (9, 0.5) \\
  label smoothing & 0.1 \\
  mixup & 0.8 \\
  cutmix  & 1.0 \\
  drop path rate & 0.1 \\
  fine-tuning resolution & 224 $\times$ 224 \\
  \shline
  \end{tabular}
  \end{table}

  \noindent\textbf{Object Detection and Instance Segmentation.}
  The COCO fine-tuning configuration is provided in Table~\ref{tab:impl_coco}. We adopt the ViT-adapted Mask R-CNN~\cite{he2017mask} with FPN~\cite{lin2017feature} following standard practice.
  
  \begin{table}[h]
  \centering
  \caption{{Fine-tuning settings for object detection and instance segmentation.}}
  \label{tab:impl_coco} 
  \begin{tabular}{y{106}|x{98}}
  \shline
  config & ViT-B \\
  \shline
  optimizer & AdamW \\
  base learning rate & 3e-3 \\
  weight decay & 0.05 \\
  optimizer momentum & $\beta_1, \beta_2{=}$ 0.9, 0.999 \\
  layer-wise decay & 0.75 \\
  batch size & 16 \\
  learning rate schedule & step decay \\
  training epochs & 12 \\
  drop path & 0.2 \\
  \shline
  \end{tabular}
  \end{table}
  
  \noindent\textbf{Semantic Segmentation.}
  The ADE20K fine-tuning configuration is summarized in Table~\ref{tab:impl_ade20k}. We use the UperNet~\cite{xiao2018unified} framework.
  
  \begin{table}[h]
  \centering
  \caption{{Fine-tuning settings for semantic segmentation.}}
  \label{tab:impl_ade20k} 
  \begin{tabular}{y{106}|x{98}}
  \shline
  config & ViT-B \\
  \shline
  optimizer & AdamW \\
  base learning rate & 4e-4 \\
  weight decay & 0.05 \\
  optimizer momentum & $\beta_1, \beta_2{=}$ 0.9, 0.999 \\
  layer-wise decay & 0.65 \\
  batch size & 16 \\
  learning rate schedule & linear decay \\
  training iterations & $160k$ \\
  drop path & 0.1 \\
  \shline
  \end{tabular}
  \end{table}
 }

\bibliography{ref}
\bibliographystyle{IEEEtran}

\vfill